\documentclass[sigconf,nonacm,natbib=false]{acmart}

\usepackage{booktabs}
\usepackage{amsmath}

\usepackage{algorithmic}
\usepackage[ruled,linesnumbered]{algorithm2e}
\usepackage{graphicx}
\usepackage{textcomp}
\usepackage{caption}
\usepackage{subcaption}
\usepackage{xcolor}
\usepackage{array,booktabs}
\usepackage{siunitx}
\usepackage{listings}
\usepackage{hyperref}

\graphicspath{{./}}

\def\figurename{Figure}
\def\tablename{Table}
\def\sectionname{Section}

\AtBeginDocument{%
  }

\definecolor{delim}{RGB}{20,105,176}
\definecolor{numb}{RGB}{106, 109, 32}
\definecolor{string}{rgb}{0.64,0.08,0.08}

\lstdefinelanguage{json}{
    numbers=none, 
    numberstyle=\small,
    frame=single,
    rulecolor=\color{black},
    showspaces=false,
    showtabs=false,
    breaklines=true,
    tabsize=1,
    postbreak=\raisebox{0ex}[0ex][0ex]{\ensuremath{\color{gray}\hookrightarrow\space}},
    breakatwhitespace=true,
    basicstyle=\footnotesize\ttfamily,
    upquote=true,
    morestring=[b]",
    stringstyle=\color{string},
    literate=
     *{0}{{{\color{numb}0}}}{1}
      {1}{{{\color{numb}1}}}{1}
      {2}{{{\color{numb}2}}}{1}
      {3}{{{\color{numb}3}}}{1}
      {4}{{{\color{numb}4}}}{1}
      {5}{{{\color{numb}5}}}{1}
      {6}{{{\color{numb}6}}}{1}
      {7}{{{\color{numb}7}}}{1}
      {8}{{{\color{numb}8}}}{1}
      {9}{{{\color{numb}9}}}{1}
      {\{}{{{\color{delim}{\{}}}}{1}
      {\}}{{{\color{delim}{\}}}}}{1}
      {[}{{{\color{delim}{[}}}}{1}
      {]}{{{\color{delim}{]}}}}{1},
}
\definecolor{commentColor}{HTML}{6A9F3D}
\newcommand{\tbf}[2][commentColor]{}

\newcommand{\revision}[1]{#1}

\setcopyright{none}

\acmSubmissionID{5144}

\RequirePackage[
  datamodel=acmdatamodel,
  style=acmnumeric,
  ]{biblatex}

\begin{document}

\title{A Multi-Agent LLM Framework for Design Space Exploration in Autonomous Driving Systems}
\titlenote{This is the extended version of the paper accepted to the 41st ACM/SIGAPP Symposium on Applied Computing. The official version is available at \url{https://doi.org/10.1145/3748522.3779714}}

\author{Po-An Shih}
\orcid{0009-0000-8964-9006}
\affiliation{%
  \institution{National Cheng Kung University}
  \city{Tainan}
  \country{Taiwan}
}

\author{Shao-Hua Wang}
\orcid{0000-0003-1961-4304}
\affiliation{%
  \institution{National Cheng Kung University}
  \city{Tainan} 
  \country{Taiwan}
}

\author{Yung-Che Li}
\orcid{0009-0007-8635-1483}
\affiliation{%
  \institution{National Cheng Kung University}
  \city{Tainan} 
  \country{Taiwan}
}
\author{Chia-Heng Tu}
\orcid{0000-0001-8967-1385}
\affiliation{%
  \institution{National Cheng Kung University}
  \city{Tainan}
  \country{Taiwan}
}
\email{chiaheng@ncku.edu.tw}

\author{Chih-Han Chang}
\orcid{0009-0009-3662-7976}
\affiliation{%
  \institution{Safeware Technology Inc.}
  \city{Taipei}
  \country{Taiwan}
}
\email{chihhan@ieee.org}




\renewcommand{\shortauthors}{P. A. Shih et al.}

\begin{abstract}
  Designing autonomous driving systems requires efficient exploration of large hardware/software configuration spaces under diverse environmental conditions, e.g., with varying traffic, weather, and road layouts. Traditional design space exploration (DSE) approaches struggle with multi-modal execution outputs and complex performance trade-offs, and often require human involvement to assess correctness based on execution outputs. This paper presents a multi-agent, large language model (LLM)-based DSE framework, which integrates multi-modal reasoning with 3D simulation and profiling tools to automate the interpretation of execution outputs and guide the exploration of system designs. Specialized LLM agents are leveraged to handle user input interpretation, design point generation, execution orchestration, and analysis of both visual and textual execution outputs, which enables identification of potential bottlenecks without human intervention. A prototype implementation is developed and evaluated on a robotaxi case study (an SAE Level 4 autonomous driving application). Compared with a genetic algorithm baseline, the proposed framework identifies more Pareto-optimal, cost-efficient solutions with reduced navigation time under the same exploration budget. Experimental results also demonstrate the efficiency of the adoption of the LLM-based approach for DSE. We believe that this framework paves the way to the design automation of autonomous driving systems.
\end{abstract}



\maketitle


\section{Introduction} \label{introduction}

Autonomous driving systems have transformed the landscape of modern transportation as they have rapidly evolved from research prototypes to commercially viable solutions. They cover a wide range of applications, from highway autopilot to advanced SAE Level 4 deployments~\cite{SAEJ3016_202104}, such as robotaxi services and automated valet parking. SAE Level 4 applications can operate without human intervention in well-defined operational environments. Achieving this level of autonomy require complex decision-making capabilities and precise vehicle control under diverse environmental conditions, which imposes strict requirements on both software performance and system reliability. Designing such systems calls for efficient exploration of vast hardware/software configuration spaces to meet application-specific constraints while minimizing hardware cost.


The design of autonomous driving systems presents substantial challenges due to the combined complexity of software organization and the variability of operational environments. Modern autonomous vehicles integrate numerous software modules, e.g., perception, planning, and control, each with interdependent functionalities and parameters affecting system performance. Furthermore, environmental variability, e.g., dynamic traffic conditions, weather conditions, and map variations, add uncertainty to the design process. These factors expand the design space to a scale where exhaustive exploration is infeasible. This situation makes efficient design space exploration a critical requirement.

Design space exploration is a systematic process used to evaluate and identify optimal hardware and software configurations for a system, with the considerations of multiple design parameters and objectives. 
Traditional DSE algorithms, such as evolutionary algorithms and machine learning models, often face difficulties in adapting to complex, multi-modal performance data generated from executions and high-dimensional configuration spaces without incurring high computational costs. Moreover, conventional DSE workflows depend heavily on manual inspection of execution outputs to verify functional correctness, such as confirming whether a vehicle’s trajectory is accurate. This manual validation is both time-consuming and prone to errors, and significantly limits the scalability and efficiency of DSE in autonomous driving systems. Therefore, developing intelligent and automated DSE techniques is critical to efficiently explore design options that can handle the uncertainties of autonomous driving environments. More about the challenges and limitations of existing works are discussed in \sectionname~\ref{sec:motivation}.

To address these challenges, we propose an LLM-based DSE framework that leverages multi-modal LLMs to automate the interpretation of execution results and guide the search for optimal configurations. The framework incorporates multiple LLM agents, each specialized in different aspects of the design space exploration process, such as one agent for taking user inputs to specify a DSE task, another for exploring the design space formed by the input design parameters, one for analyzing visual execution outputs and one for textual execution results, and another one for exercising the system execution for a predicted design point (via 3D simulations). With the inputs provided by users, this framework enables a feedback-driven exploration process that can intelligently correlate design parameters with system-level performance goals without human intervention. Our framework is prototyped as a multi-agent system, and has been applied in a case study on the robotaxi application (an SAE Level 4 application for moving from point A to point B). The experimental results show that our LLM-based approach outperforms a genetic algorithm baseline under the same exploration budget by discovering more Pareto front solutions. The results demonstrate that multi-modal LLM reasoning, together with targeted prompting strategies, can efficiently navigate large and complex design spaces. The contributions of this work are summarized as follows.
\begin{itemize}
    \item An LLM-based DSE framework for autonomous driving systems is proposed, and it integrates a multi-agent LLM architecture with established 3D simulation and profiling tools to automate the design process. To the best of our knowledge, this is the first work that applies multi-agent LLMs to the design space exploration in autonomous driving systems, leveraging feedback from the performance profiling tools to guide the exploration process. 
    \item A coordinated set of LLM agents is adopted to interpret the execution results and guide the DSE process through multi-modal reasoning over both visual and textual simulation results. This capability is the key to automating DSE tasks by enabling performance analysis and the identification of potential bottlenecks.
    \item A series of experiments have been conducted with our prototype framework, including a case study on a robotaxi application (an SAE Level 4 use case). The results demonstrate that the proposed framework discovers more cost-efficient design points compared to a genetic algorithm baseline under the same exploration budget.
\end{itemize}

The remainder of this paper is organized as follows. \sectionname~\ref{background and motivation} briefly introduces autonomous driving software and the current status of using LLMs for system design, and the motivation of this work. \sectionname~\ref{sec:framework} presents the architecture of the proposed LLM-based DSE framework, elaborates the multi-agent design, and highlights the roles of each LLM agent. \sectionname~\ref{sec:method} delineates the internals of the key components in our framework using the case study as an example. \sectionname~\ref{sec:result} provides a comprehensive evaluation, including experimental setup, experimental results on the robotaxi application, and comparative analyses against a genetic algorithm baseline and different prompting strategies. Finally, \sectionname~\ref{conclusion} concludes the work.

\section{Background and Motivation} \label{background and motivation}
\subsection{LLMs for System Design} \label{sec:LLMSD}
Recent works have explored various frameworks that integrate large language models to improve agent design and design space exploration. AgentSquare~\cite{agentsquare} presents a modular framework with standardized modules, including Planning, Reasoning, Tool Use, and Memory, enhanced through module evolution and recombination. It achieves an average performance improvement of 17.2\% over human-crafted agents, demonstrating its cost efficiency. LASER~\cite{laser} applies LLMs to web navigation by adopting a state-space exploration strategy with state-specific actions, effectively reducing errors and improving execution efficiency. Intelligent4DSE~\cite{xu2025intelligent4dse} accelerates high-level synthesis DSE by combining graph neural networks with LLM-augmented evolutionary algorithms, yielding up to 3.4× accuracy improvements in predicting latency and resource usage while eliminating expert-crafted heuristics. IDEA~\cite{chen2025idea} further augments DSE through LLM-driven constraint generation and Monte Carlo Tree Search, offering a domain-agnostic approach to systematic design exploration. It is important to note that IDEA is orthogonal to our approach as it is domain-agnostic DSE approach while our work focuses on autonomous driving system designs.

\subsection{Autonomous Driving Software} \label{sec:ads}

Autonomous driving systems like Autoware.Auto~\cite{autoware} are built on top of ROS 2 software framework~\cite{ros2}. ROS 2 software provides a flexible platform for building robotic applications, including autonomous vehicles. Each ROS software module is defined and run as a ROS node, and the ROS software employs a publish-subscribe model for inter-node communications, where \emph{message topics} serve as named channels used by publishing nodes to send messages in specific \emph{topics} and by subscribing nodes to receive messages with subscribed \emph{topics}. 
As a notable example of ROS-based systems, Autoware.Auto established by Autoware foundation is designed to support modular development and scalable deployment of autonomous vehicle systems. The autonomous driving software stack is compatible with popular open-source autonomous vehicle simulators, such as CARLA and LG SVL. These simulators enables the execution of the entire software stack of an autonomous driving system without requiring physical vehicles.

The internal workflow of Autoware.Auto follows a five-stage pipeline: \emph{sensing, localization, perception, planning,} and \emph{control}. Sensor modules collect environmental data, which is filtered (or downsampled) and forwarded to downstream modules. Localization algorithms, e.g., Normal Distributions Transform using LiDAR, estimate the vehicle’s position. Perception modules detect surrounding objects and interpret traffic conditions. A planner module then generates a trajectory, and low-level control commands are issued to the vehicle hardware to follow the trajectory. This five-stage cycle repeats in real time, and the system’s responsiveness is critical for ensuring safety. Prolonged processing time can cause unstable vehicle behavior, such as zig-zag motion.

The modular architecture of ROS 2-based autonomous driving systems allows developers to create sophisticated applications by integrating numerous specialized software modules (ROS nodes). However, this flexibility comes with significant complexity and challenges for performance analysis. For example, a typical autonomous driving scenario involves around 80 interconnected ROS nodes, while a simplified task, e.g., automated valet parking, still engages approximately 30 nodes communicating over 50 major message topics. As these nodes interact through the publish-subscribe model, the causal relationships between publishing nodes and subscribing nodes are often implicit. Identifying which nodes contribute to performance bottlenecks requires analyzing these causal links and understanding the delay propagation through the ROS nodes. Thus, analyzing system performance is a non-trivial task that demands substantial effort and proper tools to identify performance bottlenecks and optimize the system design.



\subsection{Performance Profiling for ROS-Based Autonomous Driving Systems}

A common approach to analyzing ROS-based software systems involves inserting trace probes into software to log the timing of performed operations, and these tracing tools help provide the latencies of the performed communication/computation operation. Among various tools, LTTng~\cite{lttng} has gained popularity for its ability to trace both user-space and kernel-space activities in ROS systems. Building upon LTTng, the ROS 2 tracing tool~\cite{ros2tracing} enables fine-grained analysis, including data publication/subscription frequency and  the duration of the subscription or of the timer callback function. The ROS 2 tracing tool is enhanced to recover the causal links of a ROS 2 system to analyze the latencies of messages flowing through ROS 2 nodes.

Another key focus of ROS performance tools is measuring \emph{end-to-end latency}, which reflects the time required for data to flow from a source node to a downstream sink node. Tools, such as Autoware\_Perf, extend ROS 2 tracing to support the latency measurement of selected node chains in Autoware.Auto for capturing computation and communication delays. Casini et al~\cite{Casini2019ResponseTimeAO} propose a theoretical model to bound worst-case response times with real-time scheduling analysis. Additionally, ROS-Llama~\cite{9470451} proposes an automatic latency manager to keep the end-to-end latency of selected ROS 2 nodes within real-time bounds. 

To aid system-wide analysis, several works have introduced \emph{abstract representations} of ROS applications with graph-based models. The ROS 2 built-in \texttt{rqt\_graph} tool visualizes node links and message flows.The graph representation provides developers with an overview of the communication structure of ROS nodes. Besides, an execution flow path analysis tool~\cite{exectutionflow} provides a more detailed view of the execution flow paths, starting from a ROS node in the sensing stage to a node in the control stage. The execution flow paths allow developers to rapidly identify the critical paths and potential bottlenecks in the ROS system. 
PARD~\cite{PARD2025} takes a step further by proposing data-flow aware analyses methods on the \texttt{rqt\_graph}. For example, the \emph{CPU bound analysis} identifies ROS nodes overwhelmed by high CPU usage, which cause delays or dropped data. The \emph{data frequency bound analysis} detects mismatched input
message frequencies that limit node performance, which enables adjustments node configurations (e.g., data publishing frequency) to ensure smooth system operation. These analyses significantly reduce the complexity of performance debugging and accelerate design iterations.






\subsection{Motivation} \label{sec:motivation}

Design space exploration is the process of systematically evaluating hardware/software configurations to identify optimal system designs under given performance and cost constraints. DSE is typically an iterative process covering the following steps: configuration prediction, design point evaluation, and result validation, until the design objectives are met. 
In autonomous driving systems, DSE involves exploring parameters across autonomous driving software and underlying computing hardware in a large and complex design space. Each iteration refines system configurations based on the insights learned from profiling data and application-specific objectives. Automating this iterative process is critical for accelerating the system design. The following paragraphs describe the challenges faced by conventional methods for the system design and highlight the insights of our proposed framework to address these challenges. 

\emph{\textbf{Software Complexity.}}
The modular design and publish-subscribe architecture of ROS-based autonomous driving systems introduce significant complexity, as introduced in \sectionname~\ref{sec:ads}. For instance, Autoware.Auto integrates over 80 interdependent software modules communicating via publish-subscribe mechanisms, and minor delays or inefficiencies in a single module can propagate throughout the five-stage execution pipeline, leading to critical issues like unstable or unsafe vehicle trajectories such as zig-zag trajectories. 
While performance profiling tools are available to capture runtime behaviors, interpreting the relationships between observed performance data and hardware/software design parameters remain difficult. System designers must manually analyze these performance data, identify performance bottlenecks, and predict optimal hardware configurations that ensure real-time constraints and stable vehicle behaviors while maintaining cost-efficiency (e.g., cost-efficient hardware platforms).

\emph{\textbf{Environmental Variability.}}
Autonomous vehicles must operate reliably across diverse and unpredictable environments, such as various traffic conditions and road infrastructures. The variability makes functional behavior validation of autonomous driving systems particularly challenging. Taking a ground truth trajectory as an example, a trajectory is commonly used to determine whether a vehicle’s behavior is correct and varies depending on the specific autonomous driving task. In a navigation task, the trajectory of moving from point A to point B depends on the road network defined in the map, whereas in an automated valet parking (AVP) task, the trajectory of  depends on the layout of the parking space. The complexity is further amplified by dynamic traffic conditions, which introduce variability even within the same environment. As a result, determining whether a vehicle’s behavior is \emph{functionally correct} cannot rely on a static reference but instead requires human intelligence tailored to each unique scenario. 

The above context-dependent nature of behavior validation renders traditional DSE methods unsuitable for automatically handling the design space exploration of autonomous driving systems. Traditional approaches rely heavily on manual inspection or static reference data to assess the correctness of system behaviors. When the task goal, map, or environment changes, new ground truth data must be generated for each case, along with its corresponding validation criteria. Taking the setup of our framework as an example, the ground truth trajectory for a navigation task from point A to point B on a specific map is different from that of an AVP task on the same map. This variability imposes a significant burden on the design process, as it requires human intervention to produce the context-specific ground truth data and to assess vehicle behavior by comparing observed trajectories against these references.

\emph{\textbf{LLM-Augmented DSE for Autonomous Driving.}}
This work is inspired by the potential of LLMs to interpret multi-modal data for reasoning the textual performance data and analyzing visual trajectory plots to guide a system design process.
The reasoning capabilities of LLMs enable the identification of potential system bottlenecks and the interpretation of the root cause the performance limitations. The analysis capabilities allow LLMs to automatically verify the expected behaviors of autonomous driving systems for different scenarios. 
Unlike traditional automation methods that require explicit programming for each task goal, LLMs are able to adapt to new situations using pre-trained knowledge and in-context learning.
Moreover, as LLMs are capable of correlating the execution results (the textual and visual data) with system design parameters, they are able to predict optimal hardware-software configurations that meet the design objectives. 

As a pioneer work in this research area, we propose an LLM-augmented framework to automate the design space exploration process for autonomous driving systems. 
A multi-agent LLM architecture is adopted to automate the DSE process. In particular, several LLM agents are developed to accept the inputs from system designers, to predict optimal hardware-software configurations, to interpret the performance data after the evaluation of each design configuration, and to validate the correctness of the system behaviors of each configuration. The collaboration of the multiple agents is conducted by a group chat pattern, where each agent has its own role and responsibility for the DSE. 

Third-party software is integrated for the automated DSE process, which also provides the flexibility for the design of autonomous driving systems. The third-party software includes the CARLA simulator~\cite{carla} for autonomous driving simulations, the Autoware.Auto software stack~\cite{autoware} for autonomous driving, and performance profiling tools for collecting Autoware.Auto's performance data, such as ros2\_tracing~\cite{ros2tracing} and PARD~\cite{PARD2025}. The integration of these tools enables the framework to automatically set up simulation environments, execute autonomous driving tasks, collect performance data, and analyze the results without human intervention.
As autonomous vehicles transition from research prototypes to commercial products, automating the identification of optimal system configurations becomes critical and helps accelerate time-to-market.

\section{Framework Architecture} \label{sec:framework}

\begin{figure}[htb!]
\centerline{\includegraphics[width=.75\columnwidth]{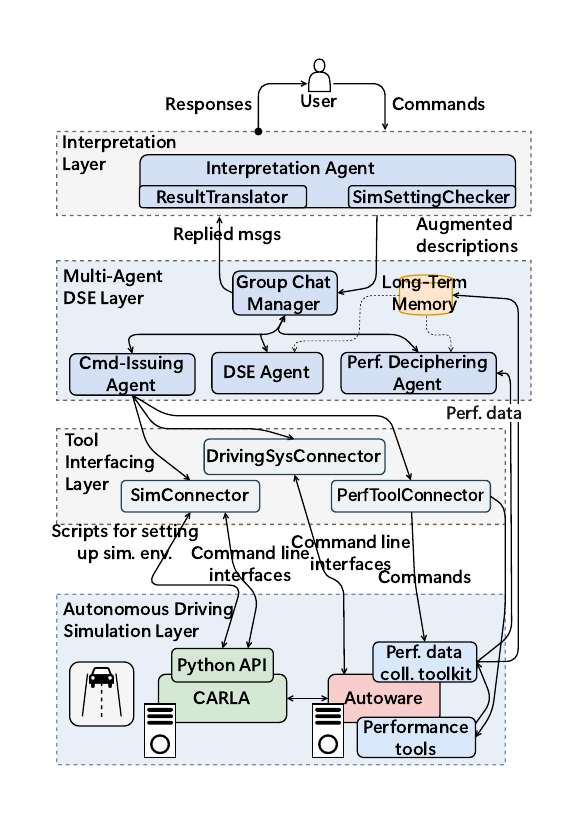}}
\caption{The architecture of the LLM-augmented DSE framework for the designs of autonomous driving systems.}
\label{fig:frameworkoverview}
\end{figure}

The LLM-augmented DSE framework consists of four layers to automate the design space exploration process for autonomous driving systems. The architecture of the framework is illustrated in \figurename~\ref{fig:frameworkoverview}. The framework follows a systematic workflow, taking user commands as input and producing optimized hardware/software configurations as output. 

The Interpretation Layer is responsible for interacting with system designers who specify high-level objectives and constraints as text. The text is then translated into actionable goals by an LLM agent. The Multi-Agent DSE Layer contains several specialized LLM agents, and they are communicate using a group-chat pattern. Each of the agents is responsible for some tasks, e.g., configuration prediction, command generation, or performance analysis and behavior validation. The core logic of the DSE process is performed within this layer. The Tool Interfacing Layer connects the LLM agents to external tools, including CARLA, Autoware.Auto, and performance profilers, to issue the generated commands. The Autonomous Driving Simulation Layer runs the simulations based on the generated commands and produces multi-modal simulation results, including textual performance data and visual vehicle trajectory. These simulation results are fed back to the Multi-Agent DSE Layer to establish a closed feedback loop for iterative DSE.

The following subsections introduce high-level functionalities for each layer. The methods used to establish the multi-agent framework are further detailed in \sectionname~\ref{sec:method}.

\subsection{Interpretation Layer} \label{sec:interpretation}

This layer enables system designers to specify system configurations and objectives for design space exploration. The high-level input is parsed by the Interpretation Agent to generate structured inputs for the design space exploration process for the downstream layer. The agent also validates input consistency and reports final DSE results. When encountering input error or inconsistency, the agent provides interactive feedback to its user for correction. The inputs include the following three categories.
\begin{itemize}
  \item Application scenarios, which define the task to be performed by the autonomous driving vehicle, e.g., lane driving or automated valet parking. In addition, the map associated with the given task and the start/goal positions of the autonomous vehicle should be provided to facilitate autonomous driving simulations.
  \item Hardware/Software configurations. Hardware configurations specify the parameters of the computing hardware for autonomous driving, such as the number of CPU cores and their operating frequencies. Software configurations refers to software settings that could be sensor data receiving rates, sensor data resolution (e.g., camera resolution) and the software version of Autoware.Auto.
  \item Design objectives, which define, for example, the constraints on execution time of a given task or cost of the computing hardware for autonomous driving. 
\end{itemize}

The framework operates under the following assumptions for the automated DSE of autonomous driving systems using 3D simulations. 
Application scenarios and hardware/software configurations are assumed to be supported by the underlying simulation environment, such as CARLA simulator and Autoware.Auto software stack. For example, the maps, vehicle models, traffic patterns, sensors (e.g., Lidar, camera, and GPS) to be simulated must be supported by CARLA. Similarly, the specified task (e.g., lane driving or AVP), sensors to be mounted on the simulated vehicle, and sensor settings must be supported by Autoware.Auto. 

Furthermore, the hardware configurations are assumed to be supported by the underlying computing hardware, and any changes to the hardware settings should be done by software scripts. For instance, the number of CPU cores and their operating frequencies should be supported by the hardware hosting Autoware.Auto. Adjusting CPU frequencies should be possible using tools, such as using \texttt{cpupower}~\cite{cpupower} in Linux.

Design objectives are assumed to be expressed in a way that can be interpreted by the LLM agents in the Multi-Agent DSE Layer. Currently, the design objectives can include constraints on execution time or cost of computing hardware, which can be interpreted by the LLM agents to guide the DSE process. When the evaluation metrics used to evaluate the design objectives cannot be \emph{reasoned} about by the LLM agents, a supporting mechanism should be provided to \emph{interpret} the simulation results and \emph{correlate} the results with the given metrics. Detailed information of the DSE problem modeling is presented in \sectionname~\ref{sec:designmodeling}.






\subsection{Multi-Agent DSE Layer} \label{sec:multiagent}

This layer is the core of the framework for handling the automated DSE process. Upon receiving the structured data from the Interpretation Layer, a Group Chat Manager in this layer orchestrates the DSE process by using a specific order of \emph{speakers} (agents) to sequentially handle the designated tasks. In particular, the process is realized by a round-robin conversation pattern. In our current implementation, the AutoGen framework~\cite{autogen} is adopted and the round-robin algorithm is supported by the framework to facilitate the DSE process. 

\emph{\textbf{Multi-Agent Based DSE Process.}}
Based on the structured inputs, a DSE Agent is first \emph{speaker} to generate an initial design point, which is a multi-dimensional data tuple containing the application scenarios and hardware/software configurations to be evaluated. This design point is then sent to a Command-Issuing Agent, which generates high-level commands to describe the actions and settings required by the external tools to evaluate the design point. These high-level commands are sent to the Tool Interfacing Layer, which generates the scripting codes to run  autonomous driving simulations and collect simulation results. The simulation results are then forwarded to a Performance Data Deciphering Agent analyzing the data and provides insights into potential performance bottlenecks. As the start of a new DSE iteration, the DSE Agent uses these insights to predict the next design point, and the above process repeats until the design objectives are met.


\emph{\textbf{Design Space Exploration Agent.}}
The DSE Agent is responsible for guiding the iterative process of searching optimal hardware-software configurations that meet user-defined design objectives. This agent takes the inputs, including design objectives, the defined design space (e.g., number of CPU cores and CPU frequency), the simulation results from a prior simulation run (where diagnostic insights of the simulation results are generated by the Performance Deciphering Agent). This agent outputs a new design point to be evaluated in the next simulation run. The design point is generated based on the reasoning about the performance bottlenecks identified in the previous simulation results and the design objectives. The reasoning about the justification of the prediction (of the new design point) is also generated by this agent for designers to reveal more about the decision making. Furthermore, in order to improve the decision-making process, a Long-Term Memory is attached to this agent to better reasoning the relationships among the design points and the collected performance data. The details of the DSE Agent are described in \sectionname~\ref{sec:dseprocess}.

\emph{\textbf{Performance Deciphering Agent.}}
This agent is responsible for analyzing the results collected from a simulation run of a given design point. It looks into both textual and visual performance data and generates insights to help the DSE Agent understand the behavioral and computational characteristics of the design point. Furthermore, as the diagnostic insights reveal potential performance bottlenecks, they help facilitate subsequent DSE iterations by predict better design alternatives (presented in \sectionname~\ref{sec:perfdatadeciphering}). 

The inputs to the Performance Deciphering Agent include application scenario (e.g., the map file, start and goal points, and the screenshot of the map with the coordinates of map corners) and multi-modal performance data collected from the simulation run (i.e., textual profiling data collected by third-party performance tools and visual vehicle trajectory screenshots obtained). Two types of LLMs, text and vision LLMs, are employed to analyze the textual and visual performance data, respectively. It is worth noting that while new LLMs are developed to integrate visual and textual data, the current framework uses two LLM instances to analyze the textual and visual data as a proof-of-concept for our framework. The potential of an integrated LLM for the DSE process can be further explored in our future work. 

A vision LLM analyzes the vehicle trajectory screenshot to assess the functional behaviors of the autonomous vehicle simulation based on checking the starting and ending points of the trajectory for this simulation run, against the data (answer) specified in the application scenario. For instance, using the trajectory screenshot, the vision LLM checks whether the simulated vehicle reaches the goal (destination) point to determine the completeness of the autonomous driving task. A positive or negative result is generated as part of the performance report to the DSE Agent. The DSE Agent marks the configuration as functionally invalid when encountering a negative result. 

A text LLM parses the profiling data collected by the performance tools (e.g., ros2\_tracing~\cite{ros2tracing}) to identify potential bottlenecks, in terms of \textit{CPU-bound issues} for CPU-overloaded ROS nodes and \textit{data-frequency-bound issues} for mismatched message transferring rates among ROS nodes~\cite{PARD2025}; these issues are detailed in the next section. Based on the empirical experiences reported in existing works~\cite{PARD2025}, performance metrics are judiciously selected as estimates to quantify the delivered performance of the autonomous driving system for a given design point (hardware/software configuration). 

The outputs of the two LLM instances are merged into a performance report, including 1) the analysis results for the moving path of the simulated vehicle, and 2) a list of potential performance bottlenecks/concerns discovered from the textual profiling data. An example of the input and output of this agent is given in the next section. This combined report is then forwarded to the DSE agent to provide intelligent feedback that helps guide the search toward the next promising design point.

\emph{\textbf{Command-Issuing Agent.}}
This agent is responsible for interpreting the design point predicted by the DSE Agent into the high-level commands that will be further converted into the executable commands to run autonomous driving simulations and performance data collections. Specifically, with the predicted design point, including hardware parameters (e.g., number of CPU cores) and software settings (e.g., LiDAR publish frequency, this agent generates three types of commands corresponding to the components in the Tool Interfacing Layer: 1) SimConnector commands to configure the autonomous driving simulation (e.g., the specified map and traffic patterns), 2) DrivingSysConnector commands to set up parameters for autonomous driving software and underlying computing hardware (e.g., adjusting CPU core numbers and specifying the lane driving task), and 3) PerfToolConnector commands for configuring performance profiling tools (e.g., profiling CPU utilizations and communication times among ROS nodes). 

\subsection{Tool Interfacing Layer}

This layer facilitates communication between the framework and the underlying autonomous driving simulation environment by sending executable commands to and receives outputs from the underlying external tools. It has three connectors, SimConnector, DrivingSysConnector, and PerfToolConnector, designed to interact with the external tools involving in the simulation environment, including CARLA simulator, Autoware.Auto software stack, and performance profiling tools. The connectors translate high-level commands from the Command-Issuing Agent into executable shell scripts or API calls to configure the simulation environment, set up and launch the autonomous driving system, and control performance profiling tools. It is important to note that while CARLA and Autoware are used as examples in the prototyping framework to illustrate its design, the proposed framework can be retargeted to other 3D simulators and ROS-based autonomous driving software by adapting these three connectors. The following paragraphs describe the functionalities of each connector, based on the assumptions listed in \sectionname~\ref{sec:interpretation}.

\emph{\textbf{SimConnector.}}
This connector helps the configuration of the 3D simulation environment based on the specified application scenario. Specifically, it issues actual commands to run the 3D simulation, such as selecting maps, choosing autonomous vehicle models, setting traffic conditions, and configuring weather parameters (e.g., clear, rain, fog). To interact with the CARLA simulator, this connector converts the high-level commands into shell commands to start the CARLA simulator and issues Python codes (via the Python APIs exposed by CARLA) to configure the simulation environment. 

\emph{\textbf{DrivingSysConnector.}}
This connector translates the hardware/software configurations into executable commands to configure the computing hardware and the autonomous driving software. 
In particular, the hardware configuration must take effect before launching the autonomous driving software. Taking our case study as an example, the adjustments to the  CPU core count and frequency are applied first by executing the corresponding commands (e.g., \texttt{taskset} and \texttt{cpufreq-set} in Linux based systems). Note that \texttt{taskset} is used to bind processes to specific CPU cores to simulate a limited-core execution environment for Autoware. The toolkit \texttt{cpufreq-set} is used to set the processor's operating frequency. 

The software configuration involves parameters within the Autoware software, such as setting up data frequencies for different sensors. In our case study, the connector changes the LiDAR subscribing frequency for the Autoware system to reflect the chosen design point. 
In addition, the information specified in the application scenario is used to configure additional Autoware parameters, such as the designated task, the map to be loaded, start and goal points, and the vehicle model. This connector incorporates this information into the executable command parameters used to launch the Autoware software.

\emph{\textbf{PerfToolConnector.}}
This connector controls the setup and execution of performance profiling tools, which collect performance data during the autonomous driving simulation. The collected data reflects the combined effects of the application scenario (e.g., automated valet parking in a specific space) and the software/hardware configurations for a specific design point. In the current implementation, the connector supports two profiling tools, ros2\_tracing~\cite{ros2tracing} and PARD~\cite{PARD2025}, which are used to collect execution traces and perform dataflow aware analyses, respectively. It is important to note that as introduced in \sectionname~\ref{sec:ads}, ros2\_tracing provides function- and message-level timing information for each ROS node, while PARD offers higher-level dataflow aware analyses among ROS nodes to identify potential performance bottlenecks in a ROS system, i.e., CPU-bound and frequency-bound issues. By using different performance tools, the framework can further assist the text LLM of the Performance Deciphering Agent in understanding the performance characteristics of the autonomous driving system and identifying potential performance bottlenecks. 

Example outputs generated by these tools are provided in the following section. The connector issues commands to start the profiling tools before launching the Autoware software, and stops the profiling tools after the simulation run is completed. The collected performance data are analyzed by the dataflow aware analyses to highlight potential performance bottlenecks and are sent to the Performance Deciphering Agent to guide the DSE process.

\subsection{Autonomous Driving Simulation Layer}

This layer implements a co-simulation environment~\cite{wang2022automatic} using two separate machines, one for mimicking a real-world environment with CARLA and the other for running the autonomous driving software with Autoware.Auto. 
In this setup, the CARLA simulator represents the environment in which the autonomous driving system operates, while the second machine simulates the computing hardware of the target vehicle. As a result, the performance observed on the second machine serves as an indicator for the delivered performance on the actual vehicle hardware.

The CARLA simulator generates virtual sensor data, such as LiDAR and cameras, to characterize the surrounding environment, while the Autoware software takes these sensor data as input and returns control signals (to CARLA), mimicking a real-world vehicle control loop. The interaction between the simulator and the autonomous driving software allows for comprehensive testing of the autonomous driving vehicle's functional behavior under various conditions specified in the application scenario with different hardware/software configurations. 
The profiling tools are also integrated into the simulation environment to collect performance data during the execution of autonomous driving tasks. 
The outputs of this layer, including vehicle trajectories and collected performance data (e.g., the latencies of ROS messages and the execution time for performing the designated task), are crucial for validating whether the vehicle's behavior is correct and meets predefined objectives. The details of the collected performance data are described in the next section.



\section{Methodology} \label{sec:method}

The core logic of the DSE process is introduced in this section.
\sectionname~\ref{sec:designmodeling} presents the DSE problem modeling, which provides the foundation for the case study on autonomous driving system design established in this work. The case study is used to illustrate the overall DSE process, including user input for the DSE process (\sectionname~\ref{sec:userinput}), the DSE procedure itself (\sectionname~\ref{sec:userinput}), and the reasoning applied to the performance data (\sectionname~\ref{sec:perfdeciphering}). In particular, each subsection describes the design of the corresponding LLM agent. 

\subsection{DSE Problem Modeling} \label{sec:designmodeling}
A design space exploration task is defined by application scenarios, hardware/software configurations, and design objectives. While system designers are responsible for defining the DSE task, our framework helps expedite the problem definition and the search process. In particular, in the current prototype of the framework, we assist system designers in specifying the application scenarios, as well as design objectives, as they are bounded by the assumptions described in \sectionname~\ref{sec:interpretation}. System designers are able to explore the design space by specifying desired hardware/software configurations. It is important to note that system designers are also allowed to adjust the application scenario and design objectives, as long as these are supported by the underlying assumptions.

As the goal is to support DSE of SAE Level 4 driving automation, the application scenarios supported by this prototyping framework include any autonomous driving tasks for moving the vehicle from point A to point B. Examples include self-driving taxi (referred to as the lane driving in this work) and automated valet parking tasks. 
\revision{For hardware/software configurations, Autoware.Auto on top of ROS~2 in principle exposes a rich set of tunable parameters, including CPU-related knobs (number of active cores, per-core frequency), GPU power limits, QoS profiles, and sensor-level settings such as LiDAR publishing frequency and detection range, as well as camera resolution and planner variants in the behavior-planning stack. In the current prototype, however, the explored design space is intentionally constrained to three parameters: number of CPU cores, CPU core frequency, and LiDAR data subscription frequency in order to keep the search space small enough for exhaustive enumeration and fair comparison against the baseline methods, and focus on demonstrating that the proposed multi-agent LLM framework can close the DSE loop. Extending the connector layer to expose additional GPU-, memory-, and planner-related parameters only requires modifying the command templates in the DrivingSysConnector, and is therefore orthogonal to the multi-agent orchestration mechanism evaluated in this work.}

With these application scenarios in mind and based on the empirical insights reported in PARD~\cite{PARD2025}, the metrics used to evaluate the execution efficiency of the autonomous driving tasks are defined as follows. 
First, the \emph{navigation time} ($T_{\mathit{Nav}}$) is the total time taken by the autonomous vehicle to complete the task, which is measured from the start point to the goal point. Second, the \emph{trajectory deviation score} ($S_{Dev}$) is a normalized distance between the ideal trajectory and the actual trajectory. The ideal trajectory can be obtained before the DSE process to served as a reference (i.e., ground truth) to quantitatively measure the quality of the vehicle trajectory simulated under a given hardware/software configuration. Third, the \emph{control command issue rate} ($\mathit{Ctrl}_{freq}$) is the frequency at which control commands (e.g., steering, acceleration, and braking) are published by the autonomous driving software. This rate evaluates the responsiveness of the autonomous driving system, and is critical for ensuring safe and smooth driving behavior~\cite{Becker2020}.

The design objectives are to suggest cost-efficient hardware-software configurations that  minimize navigation time and trajectory deviation score while maximizing control command issue rate. \revision{These three metrics are chosen because together they capture the critical aspects of an SAE Level~4 robotaxi scenario from both the passenger and system perspectives. For example, navigation time directly reflects the efficiency of a mission like how quickly the vehicle can complete a given trip, which is a quality-of-service objective in mobility-on-demand applications. Furthermore, the trajectory deviation score represents how closely the executed path follows a planned trajectory, serving as a proxy for lateral stability meaning that large deviations are typically associated with zig-zag motion or aggressive corrections that would be unacceptable in real deployments. Last, the control command issue rate measures the responsiveness of the low-level control loop by quantifying how frequently Autoware publishes \texttt{/control/command/control\_cmd} messages, which directly affects the ability of the vehicle to react to changes in the environment and maintain smooth actuation.} In particular, the hardware configurations refer to the parameters of the computing hardware that runs the autonomous driving system, while the software configurations refer to the parameters of the autonomous driving software. The design space is defined by these hardware/software configurations, which are further detailed in \sectionname~\ref{sec:userinput}.

\subsection{User Inputs} \label{sec:userinput}
The user inputs for a DSE process are specified in nature language, which includes an application scenario, hardware/software configurations, and design objectives. The inputs are provided as text descriptions, which are parsed by the Interpretation Agent to generate structured inputs for the DSE process. 
The Interpretation Agent is first trained with a System Prompt, which defines its role and responsibilities in the DSE process. In the current implementation, the system prompt contains the information described in \sectionname~\ref{sec:designmodeling} to define a DSE task. Besides, in order to ensure the correctness of the user inputs, the system prompt will be provided with the complete list of application scenarios, hardware/software configurations, and design objectives that are supported by the framework. The agent will prompt the user for clarification if the user inputs do not match the supported scenarios or configurations.

Guided by the system prompt, a system designer can specify a DSE task by providing input to the Interpretation Agent, which parses the desired hardware/software configurations and generates structured data for the DSE Agent. An example user input is shown in \figurename~\ref{fig:i-interpreter}. In this example, a simplified version of the user input is shown, covering only the application scenario and design objectives. As the hardware/software configurations are absent, the Interpretation Agent uses all available parameters as the target hardware/software configuration. A default traffic and weather condition is applied in the simulation. The example output of structured data sent to the DSE Agent can be found in \figurename~\ref{fig:s-dse}. When the optimal configurations are identified by the DSE Agent, they will be reported by the Interpretation Agent to the user. An example of an optimal/suggested design point is shown in \figurename~\ref{fig:o-dse}.


\begin{figure}[htb!]
\begin{lstlisting}[
  language=Python,
  literate={'}{{'}}1 {"}{{\textquotedbl}}1,
  basicstyle=\ttfamily\scriptsize,
  columns=fullflexible,
  keepspaces=true,
  breaklines=true,
  breakatwhitespace=false,
  aboveskip=2pt,
  belowskip=2pt,
  showstringspaces=false
]
$I \rightarrow$
I want to simulate a lane driving task and use the parkingslot map in simulation. The start point is (x: -4.97, y: -20.33) and goal point is (x: -9.47, y: -1.47), and then I would like the navigation time to be under 400 seconds. The simulation should also achieve a low trajectory deviation score and maintain efficient control command topic rates.

\end{lstlisting}
\caption{The user instruction example for the Interpretation Agent to start a DSE task.}
\label{fig:i-interpreter}
\end{figure}





\subsection{LLM-Guided DSE} \label{sec:dseprocess}
The DSE Agent takes the structured outputs from the Interpretation Agent and performs the design space exploration process. It is the core reasoning component responsible for searching the design space to identify optimal hardware/software configurations that meet the given objectives. The structured inputs to this agent include: 1) the application scenario, 2) the design objectives, 3) the design space defined by the given hardware/software configurations, 4) performance analysis produced by the Performance Deciphering Agent, and 5) historical performance data associated with design point information, which is offered by the Long-Term Memory module and is omitted in the following example. The fourth and fifth items are updated after each simulation run for the agent to make context-aware configuration predictions; these items are used in the second-stage prompting mechanism as described below. 

\revision{In the current prototype, the Long-Term Memory attached to the DSE Agent is implemented as an append-only collection of plain-text records rather than a vector-database-based retrieval system. After each simulation run, the framework logs the evaluated design point together with the resulting metrics and the summarized bottleneck annotations from the Performance Deciphering Agent into a structured text file. For the next iteration, the Dynamic Instruction passed to the DSE Agent includes both the latest result and a curated subset of these historical records (e.g., several representative high-cost/low-cost and high-/low-performance configurations), loaded via the agents tool interface.}

\revision{We deliberately avoid using a generic semantic-retrieval (RAG) pipeline on profiling traces because most of the performance data are numeric time-series and topic frequencies with limited semantic structure; preliminary experiments showed that vector-based retrieval tends to bias toward arbitrary semantically similar snippets rather than the most informative operating points in terms of metrics. Instead, the current design follows a configuration-aware logging (CAG-style) approach, where the DSE Agent receives explicitly selected historical configurations and their quantitative outcomes, enabling it to reason over concrete metric trends rather than approximated embeddings.}

The DSE Agent is driven by a two-stage prompting mechanism. The first stage uses a System Prompt (S), which encodes the agent’s role and high-level exploration goals. For example, in the current implementation, the objective is to minimize navigation time and trajectory deviation score while maximizing control command issue rates. The design space includes three configurable parameters: the number of CPU cores, CPU core frequency, and LiDAR data subscription frequency. The second stage prompting is done by using \emph{Dynamic Instructions} (I), which are constructed for each DSE iteration. Each prompt I includes the information of the last design point (i.e., the configuration of the last design point, the associated performance data, and the associated analysis report from the Performance Deciphering Agent), and the historical performance data (maintained by the Long-Term Memory module). Based on this input, the agent proposes a new configuration and explains its rationale. 

Upon receiving $S$, the DSE Agent generates the initial predicted design point, and delegates the Command-Issuing Agent to produce the corresponding simulation instructions. When the simulation is completed, the DSE Agent constructs the dynamic instruction $I$ based on the simulation results reported by the Performance Deciphering Agent, and generates the next predicted configuration (as the output $O$) based on the given $I$. The process of dynamic instruction construction and configuration prediction is repeated until an optimal configuration is found. Examples of the system prompt and dynamic instructions are shown in \figurename~\ref{fig:s-dse} and \figurename~\ref{fig:i-dse}, respectively. Note that in the preliminary study, the application scenario is fixed (e.g., either a  lane driving or AVP task under fixed traffic and weather conditions), and the design space is defined by three parameters: the number of CPU cores, CPU core frequency, and LiDAR data subscription frequency. The design objectives are to minimize navigation time and trajectory deviation score while maximizing control command issue rates.

\begin{figure}[htb!]
\begin{lstlisting}[
  language=Python,
  literate={'}{{'}}1 {"}{{\textquotedbl}}1,
  basicstyle=\ttfamily\scriptsize,
  columns=fullflexible,
  keepspaces=true,
  breaklines=true,
  breakatwhitespace=false,
  aboveskip=2pt,
  belowskip=2pt,
  showstringspaces=false
]
$S \rightarrow$
You are a design space exploration agent.
Your job is to search the design space and recommend hardware/software configurations that meet the design objectives. The objectives are:
- Minimize navigation time
- Minimize vehicle trajectory deviation score 
- Maximize control command issue rates
The application scenario is fixed to a lane driving scenario on the OpenSkyParkingLot map.

The design space is defined by three design parameters as follows:
- number_of_cores $\in$ {1 .. 28}
- core_frequency (GHz) $\in$ {1.0, 1.2, 1.5, 1.8, 2.1}
- LiDAR data subscription frequency (Hz) $\in$ {7, 14}

Your inputs will be a structured message containing:
(i) The predicted design point (hw/sw configurations) from the previous DSE iteration.
(ii) The associated performance data with the previous predicted design point.
(iii) The analysis generated by the Performance Deciphering Agent (previous agent).

Your output is a design point expressed as tuples of the form (number_of_cores, core_frequency, LiDAR_frequency), and your explanation of your predicted design point.
\end{lstlisting}
\caption{The example system prompt, $S$, of our case study.}
\label{fig:s-dse}
\end{figure}

\begin{figure}[htb!]
\begin{lstlisting}[
  language=Python,
  literate={'}{{'}}1 {"}{{\textquotedbl}}1,
  basicstyle=\ttfamily\scriptsize,
  columns=fullflexible,
  keepspaces=true,
  breaklines=true,
  breakatwhitespace=false,
  aboveskip=2pt,
  belowskip=2pt,
  showstringspaces=false,
  commentstyle=\color{black}
]
$I \rightarrow$
Here are some design points you can refer:
# Reference design point 1:
number_of_cores = 6, core_frequency = 1.5, lidar_frequency = 7 $\rightarrow$ navigation_time = 505.43, car_trajecotory_normalized_score = 0.002868, control_command_issue_rates = 2.021, hardware_cost = 9

# Reference design point 2:
number_of_cores = 23, core_frequency = 1.5, lidar_frequency = 7 $\rightarrow$ navigation_time = 171.13, car_trajecotory_normalized_score = 0.002677, control_command_issue_rates = 7.11, hardware_cost = 34.5

# Reference design point 3:
number_of_cores = 18, core_frequency = 1.8, lidar_frequency = 7 $\rightarrow$ navigation_time = 165.04, car_trajecotory_normalized_score = 0.002615, control_command_issue_rates = 7.838, hardware_cost = 32.4

Please predict the next design point for the next simulation and provide your reasoning.
\end{lstlisting}
\caption{The example of dynamic instruction before a simulation, $I$, offering a few randomly-chosen design points to refer.}
\label{fig:first-i-dse}
\end{figure}

\begin{figure}[htb!]
\begin{lstlisting}[
  language=Python,
  literate={'}{{'}}1 {"}{{\textquotedbl}}1,
  basicstyle=\ttfamily\scriptsize,
  columns=fullflexible,
  keepspaces=true,
  breaklines=true,
  breakatwhitespace=false,
  aboveskip=2pt,
  belowskip=2pt,
  showstringspaces=false
]
$I \rightarrow$
Here is the design point in simulation:
- number_of_cores: 4
- core_frequency: 1.2 GHz
- LiDAR frequency: 14 Hz

Here is the performance data after simulation:
- navigation time: 1248.49 seconds
- car trajecotory normalized score: 0.003092
- control command issue rates: 0.911 Hz

{Performance analysis from Deciphering Performance Data Agent}

Please predict the next design point for the next simulation and provide your reasoning.
\end{lstlisting}
\caption{The example dynamic instruction after a simulation, $I$, of our case study.}
\label{fig:i-dse}
\end{figure}


\begin{figure}[htb!]
\begin{lstlisting}[
  language=Python,
  literate={'}{{'}}1 {"}{{\textquotedbl}}1,
  basicstyle=\ttfamily\scriptsize,
  columns=fullflexible,
  keepspaces=true,
  breaklines=true,
  breakatwhitespace=false,
  aboveskip=2pt,
  belowskip=2pt,
  showstringspaces=false
]
$O \rightarrow$
Based on the profiling data, your primary bottlenecks are CPU-bound subscriber callbacks (especially in the behavior_velocity_planner chain) and an extremely low input frequency on the control_cmd topic (0.911 Hz).
Increase CPU frequency from 1.2 GHz $\rightarrow$ 1.8 GHz to accelerate segments (e.g., behavior planning) without incurring the thermal risks of 2.1 GHz.
Keeping your LiDAR at 14 Hz preserves trajectory accuracy.

Next design point: (16 cores, 1.8 GHz, 14 Hz LiDAR frequency)

With 16 x 1.8 GHz you should see:
- Navigation time: over 75% reduction (fewer callback stalls).
- Normalized trajectory score improve or stay flat (sensor rate unchanged).
- Control command issue rates increase from 0.91 Hz to over 3 Hz.
\end{lstlisting}
\caption{The example Output, $O$, generated by the DSE agent with the input, $I$, in \figurename~\ref{fig:i-dse}.}
\label{fig:o-dse}
\end{figure}

The Output (O) illustrated in \figurename~\ref{fig:o-dse} is generated by the DSE Agent for the dynamically-generated input (I) in \figurename~\ref{fig:i-dse}. The output includes the summary of the performance of the evaluated design point (number of cores: 4, core frequency: 1.2 GHz, liDAR frequency: 14 Hz), the predicted design point, and the reasoning about the prediction. The predicted design point is a tuple of hardware/software configurations (i.e., number of cores: 16, core frequency: 1.8 GHz, liDAR frequency: 14 Hz), which will be used as input for the next simulation run. The reasoning at the bottom of  \figurename~\ref{fig:o-dse} explains how the predicted design point is expected to improve performance based on the previous simulation results and performance analysis. 

\subsection{Deciphering Simulation Results} \label{sec:perfdeciphering}
Deciphering simulation results is a critical step in the DSE process to characterize the relationships between the given configuration and the delivered performance results. These results help facilitate the prediction of the next design point. As mentioned in \sectionname~\ref{sec:multiagent}, the Performance Deciphering Agent leverages two LLMs for interpreting both textual and visual data to provide insights into the vehicle's functional behavior and performance under the given design point. The agent operates after every simulation run and produces a structured report to the DSE Agent to guide the next iteration of the DSE process.

\emph{\textbf{The visual LLM}} analyzes the vehicle's trajectory captured from the visualization tool during the autonomous driving simulation, referred to as the \emph{actual trajectory}. The analysis focuses on two aspects: 1) the completion of the navigation task, and 2) the quality of the actual trajectory. 
In the current prototype, RViz serves as the 3D visualization tool in Autoware.Auto to revealing the simulation status. Since the start and goal positions are visually identifiable in RViz screenshots, the LLM can determine the functional behavior of the simulated vehicle. 

For a given DSE task, the visual LLM compares the actual trajectory generated by the simulated vehicle with the \emph{ideal trajectory} to evaluate the status of the navigation task and trajectory quality. The ideal trajectory may be manually provided by system designers or generated by the visual LLM using the map (in the Lanelet2 format providing lane information) and start/goal points specified in the DSE task. The current implementation adopts the latter method to automatically produce a reference path under an ideal traffic and weather conditions. For a sophisticated surrounding environment, the ideal trajectory can be generated by the autonomous driving system running under a high-performance hardware/software configuration. This is assumed to be the best configuration for the given DSE task and system designers should validate if the ideal trajectory is an acceptable reference path. 

\emph{The status of a navigation task} is determined by examining whether the vehicle has reached its goal position, given the captured actual trajectory. If so, the status of the simulated vehicle is marked as ``Navigation Completed.'' Otherwise, it is reported as ``Navigation Incomplete.'' The vehicle status is included as part of the performance report sent to the DSE Agent, along with the trajectory quality.

\emph{The trajectory quality} consists of the quantitative measurements and qualitative descriptions of the actual trajectory. The quantitative measurement is the \emph{trajectory deviation score}, which is a normalized absolute distance between the actual and the ideal trajectories. The absolute difference is computed as the Euclidean distance between corresponding points, and normalized to a range between zero and one. 
The quality of the trajectory is assessed by the visual LLM based on visual characteristics of the trajectory, such as the smoothness of the trajectory, whether it follows the lane centerline, and whether it deviates significantly from the ideal trajectory. An example of the System Prompt, $S$, of the visual LLM is illustrated in \figurename~\ref{fig:s-visualllm}, and the Output, $O$, of this prompt after analyzing the vehicle trajectory is provided in \figurename~\ref{fig:o-perfdeciphering} as part of the performance report sent to the DSE Agent. 

\begin{figure}[htb!]
\begin{lstlisting}[
  language=Python,
  literate={'}{{'}}1 {"}{{\textquotedbl}}1,
  basicstyle=\ttfamily\scriptsize,
  columns=fullflexible,
  keepspaces=true,
  breaklines=true,
  breakatwhitespace=false,
  aboveskip=2pt,
  belowskip=2pt,
  showstringspaces=false
]
$S \rightarrow$
You are a visual LLM for Performance Deciphering Agent
Your job is to use `load_image` to load car trajectory screenshot and analyze the screenshot with following descriptions.
The goal point consists of a blue dot with red and green lines forming a right-angle marker.

The application scenario are a lane driving task and the car trajectory is shown in light blue line.
The coordinates of screenshot's upper left corner and lower right corner are (x: -20.973, y: 35.428) and (x: 87.527, y: -56.072) respectively on the OpenSkyParkingLot map.
Besides, the coordinates of start point and goal point are (x: -4.973, y: 20.328) and (x: -9.470, y: -1.472) respectively on the OpenSkyParkingLot map.
The attachment is a screenshot of RViz after the Autoware simulation is completed or timeout. There may be cases where the navigation is not finished (due to timeout).

First locate the car's current position and determine:
1. Did the car reach at the goal point? If yes, then use `determine_deviation_score` to determine the deviation score based on the planned path. If the car did not reach the goal point, do not need to compute the deviation score.
2. Whether the car trajectory has significant deviation (like zig-zag), jitter or not? or whether crossed or almost crossed the adjacent lane?
3. How does the car trajectory look like?
\end{lstlisting}
\caption{The example system prompt, $S$, for the visual LLM of the Performance Deciphering Agent in our case study.}
\label{fig:s-visualllm}
\end{figure}

\emph{\textbf{The text LLM}} analyzes the textual profiling data collected by external performance tools during simulation. The raw data returned by these performance tools are converted into ROS 2 level program behaviors. This higher level performance data representation can better reflect performance issues (i.e., CPU- and frequency-bound issues) and facilitate the decision-making of the DSE task. 

In the prototype framework, ros2\_tracing is used to collect the execution traces during the system execution, where each trace record includes the timestamp of the record, ROS node name, the invoked message transmitting function, message publish rates, thread identifier (TID), and other performance logs. 
The text LLM is prompted to convert the raw data from a binary format (Common Trace Format) into a human-readable text format, which is done by leveraging an open-source tool, babeltrace. Then, the LLM converts the textual performance data into the ROS 2 level performance profiles, such as message topic publish frequency, and callback latency. The LLM also helps detect potential system issues, i.e., CPU-bound or frequency-bound behavior, from within the ROS 2 level data. The analyzed results are returned as a structured JSON summary. An example system prompt for the text LLM is shown in \figurename~\ref{fig:s-textllm}.

Particularly, the LLM helps extract the frequency of the control commands (in ROS messages) issued by the autonomous driving software. As mentioned in \sectionname~\ref{sec:designmodeling}, the control command issue rate is a critical metric reflecting the responsiveness of an autonomous driving vehicle. This rate can be obtained by analyzing the frequency of the control command topic ('/control/command/control\_cmd') in the converted ROS 2 level performance profiles. 

Besides, the LLM detects potential performance issues by analyzing the latencies of the invoked functions and message publishing rates to identify potential CPU-bound or frequency-bound issue~\cite{PARD2025}. For example, if the callback latencies of a node are significantly higher than expected, it may suggest that the node is too busy to handle the computations (a potential CPU-bound issue). Similarly, if the publishing rate of a message topic is lower than expected, it may suggest that the system has a potential frequency-bound issue. 
By default, the text LLM uses its internal knowledge to reason about the performance issues. As shown in \figurename~\ref{fig:s-textllm}, the LLM is asked to detect such issues by itself. However, it is possible to provide the LLM with the performance issue identification knowledge (e.g., the above descriptions) to help it reason about the performance issues. This can be achieved by attaching the knowledge to the system prompt shown in \figurename~\ref{fig:s-textllm}. With the reasoning capability for identifying the performance issues, the DSE process can be more efficient, as the DSE Agent can make better decisions based on the performance analysis results to shorten the design space exploration process. The experimental results in \sectionname~\ref{sec:result} further analyze the impact of the performance issue identification on the DSE process.

The LLM also helps extract the navigation time from the performance data. The autonomous driving software, Autoware.Auto, provides a Python script to connect to its execution engine, so that external users are allowed to start and stop an autonomous driving task via the script. As part of the performance data reported by profiling tools, we have added logging code to track the status of the simulated vehicle, and the navigation time is the time difference between the occurrence of the start event and the event of reaching the goal point. 

The output of the Performance Deciphering Agent aggregates the results from both the visual and text LLMs, including trajectory status, trajectory deviation score ($S_{Dev}$), navigation time ($T_{\mathit{Nav}}$), control command issue rate ($\mathit{Ctrl}_{freq}$). The data aggregation is performed by the text LLM, and an example of aggregated output is shown in \figurename~\ref{fig:o-perfdeciphering}.

\begin{figure}[htb!]
\begin{lstlisting}[
  language=Python,
  literate={'}{{'}}1 {"}{{\textquotedbl}}1,
  basicstyle=\ttfamily\scriptsize,
  columns=fullflexible,
  keepspaces=true,
  breaklines=true,
  breakatwhitespace=false,
  aboveskip=2pt,
  belowskip=2pt,
  showstringspaces=false
]
$S \rightarrow$
You are a text LLM for Performance Deciphering Agent.
Your job is to:
1. Combine outputs: Take the visual LLM (previous) agent's output, place it above your own output and return the combined output.
2. Profiling data analysis: Analyze the trace data to detect performance bottlenecks and system issues.

Your input will be a ROS2 trace from ros2_tracing, already converted using babeltrace. You can use `load_text_file(profiling_filename)` to read the file named "normal_output".
Performance issues analysis logic shows as follows:
Analysis Logic:
1. CPU-Bound Detection:
- Compare publisher vs subscriber callback frequencies for the same topic.
- If the subscriber's frequency is consistently lower than the publisher's which means exceeding a threshold that gets stricter as the frequency increases, then mark the node as CPU-bound (likely unable to keep up due to CPU overload).

2. Data Frequency-Bound Detection:
- For nodes with multiple input topics, check if input frequencies are inconsistent.
- If one input frequency is more than 10% lower than the others, the node's output rate may be limited by the slowest input (indicating a misconfiguration or upstream bottleneck).
\end{lstlisting}
\caption{The example system prompt, $S$, for the text LLM of the Performance Deciphering Agent in our case study.}
\label{fig:s-textllm}
\end{figure}




\begin{figure}[htb!]
\begin{lstlisting}[
  language=json,
  literate={'}{{'}}1 {"}{{\textquotedbl}}1,
  basicstyle=\ttfamily\scriptsize,
  columns=fullflexible,
  keepspaces=true,
  breaklines=true,
  breakatwhitespace=false,
  aboveskip=2pt,
  belowskip=2pt,
  showstringspaces=false
]
$O \rightarrow$
The navigation time is 165.04 seconds, and the control command issue rate is 7.838 Hz.
Visual LLM: 
(1) Yes, the car did reach the goal point (normalized deviation score: 0.003205); 
(2) Image shows no zig-zag, jitter, or erratic behavior; 
(3) Trajectory is stable/smooth/clean and remains within lane boundaries (including corners).
Text LLM: 
{"node_callback_latencies":{...,"663079":{"node_name":"ekf_localizer","avg_callback_latency_sec":0.000137,"max_callback_latency_sec":1.124935},...},
"topic_publish_rates":{...,"/localization/pose_twist_fusion_filter/pose":0.316,...},
"detected_issues":[...{"type":"frequency_bound","node_pid":663079,"slow_topic":"/localization/pose_twist_fusion_filter/pose","slow_hz":0.316,"fast_topic":"/perception/occupancy_grid_map/map_updates","fast_hz":135.008},...],
"bottleneck_flags":["frequency_bound"]}
\end{lstlisting}
\caption{The example Output, $O$, returned by the Performance Deciphering Agent in our case study.}
\label{fig:o-perfdeciphering}
\end{figure}

\section{Results} \label{sec:result}

\subsection{Experimental Setup} \label{sec:experimentsetup}
The prototype of the multi-agent based DSE framework is implemented using the AutoGen framework~\cite{autogen} for orchestrating the LLM agents. To facilitate the DSE process, the prototype framework is running on local machines. The selection of the LLMs for the involved agents mentioned in \sectionname~\ref{sec:framework} is based on the hardware requirements and the performance of the LLMs. The Interpretation Agent and the Command-Issuing Agent use the Llama 3.1 model~\cite{Llama3.1}, which is a lightweight model suitable for interpreting high-level inputs and generating simulation commands. The DSE Agent uses the DeepSeek-R1 model~\cite{DeepSeekR1}, which is a freely available LLM that can fit in our experimental environment and provide the reasoning capability to provide detailed reasoning information for the DSE decisions. The Performance Deciphering Agent uses the Mistral-small3.1 model~\cite{Mistral3.1} for analyzing visual performance data and uses the Llama 4 Scout model~\cite{Llama4scout} for analyzing long-text performance data. \revision{All LLM agents are instantiated using locally hosted models via the Ollama runtime to avoid dependency on commercial APIs and to ensure reproducible, cost-free experimentation. Concretely, the Interpretation Agent and the Command-Issuing Agent use Llama 3.1 as a general-purpose base model, as it provides stable language understanding and generation for purely textual tasks without requiring vision or explicit reasoning optimizations. The DSE Agent, which must reason over historical configurations and multi-dimensional performance signals, adopts DeepSeek-R1 as a reasoning-oriented model to improve its ability to extrapolate from previous design points and justify configuration updates. The visual part of the Performance Deciphering Agent employs Mistral-small3.1 with vision support, which was empirically selected among the vision-capable models available on Ollama at the time of experimentation due to its higher accuracy in detecting whether the vehicle reaches the goal and in qualitatively judging trajectory smoothness from RViz screenshots. On the other hand, the textual part of the Performance Deciphering Agent uses Llama 4 Scout model for its exceptionally large context window, which makes it suitable for handling long contexts and analyzing profiling data. This heterogeneous model assignment reflects a pragmatic trade-off between reasoning capability, vision support, and inference latency on the local machines, and the framework itself remains agnostic to the specific backbone models as long as they satisfy the required interfaces.}

As for the software in the autonomous driving simulation layer, the 3D simulation is performed by CARLA 0.9.13~\cite{carla}, and the simulated vehicle is controlled by Autoware.Auto (based on ROS 2 Foxy). A custom-built parking space map is used for the simulation to perform the designated navigation tasks, lane driving and automated valet parking. The co-simulation environment is used to run CARLA and Autoware.Auto together, where one physical machine is used to run the 3D simulation and the other runs the Autoware.Auto software stack to control the simulated vehicle. The performance tool, ros2\_tracing and PARD, runs along with Autoware.Auto to collect the runtime performance data, if necessary. The hardware and software specifications of the co-simulation environment used in the experiments are listed in \tablename~\ref{tab:cosimulaiton_env}, and another physical machine with the same specification as the PC 2 is used to run the AutoGen framework and the LLMs. 

\begin{table}[hbt!]\small
  \centering
    \caption{The hardware and software specifications for the co-simulation for the autonomous driving system design.}
    \label{tab:cosimulaiton_env}
  \begin{tabular}{p{1.25cm}p{3.0cm}p{3.0cm}}
    \toprule
    \textbf{} & \textbf{PC 1: 3D simulation} & \textbf{PC 2: Self driving SW} \\
    \midrule
    \textbf{Software} & Carla simulator 0.9.13, Windows 10 & Autoware.Auto 1.0.0, ROS 2 Galactic, Ubuntu Linux 20.04\\ 
    \textbf{Hardware} & AMD Ryzen 7 1700 processor (8 cores), 24GB RAM, Radeon RX480 GPU & Intel i7-14700 processor (20 cores w/ hyperthreading), 128GB RAM, NV RTX 3060 \\
    \bottomrule
  \end{tabular}
\end{table}

Based on two SAE Level 4 autonomous driving scenarios, robotaxi (lane driving) and automated valet parking, we have conducted a series of experiments to evaluate the performance of the proposed DSE framework. However, due to the paper length limitation, only the robotaxi results are presented here. 
In the following subsections, we will first exhibit the results for the DSE of a robotaxi system, compared with the exhaustive search method. We compare the search efficiency of our LLM-based approach with that of a genetic algorithm based approach under the same exploration budget. 


\subsection{Robotaxi System Design} \label{sec:perfdatadeciphering} 
A robotaxi application is used to demonstrate the performance of the LLM-based DSE framework. In the application, a robotaxi customer requests a ride from his current position to the entrance of a parking space, so that the customer can pick up his car from the parking space, and the robotaxi system is responsible for planning the route (from the current position to the parking space) and controlling the vehicle to complete the lane driving task. The DSE task is to find the optimal hardware/software configurations for the self-driving vehicle to complete the task within a given time limit on the parking space map while minimizing the hardware cost\footnote{\revision{In this case study, the hardware cost $C_{\text{HW}}$ is modeled as a simple proxy that is proportional to the aggregate CPU capacity of the computing platform, defined as}
\[
\revision{C_{\text{HW}} = N_{\text{cores}} \times f_{\text{core}},}
\]
\revision{where $N_{\text{cores}}$ denotes the number of CPU cores and $f_{\text{core}}$ denotes the per-core clock frequency. This linear model intentionally ignores non-CPU resources (e.g., GPU, memory, network interfaces) because the explored design space only varies CPU core counts, CPU frequencies, and LiDAR subscription rates, and the LiDAR rate does not change the computational capability of the simulation machine. The cost model is thus sufficient to differentiate CPU configurations within the constrained design space, while more detailed cost models that incorporate GPU and memory will be considered in future extensions of the framework.}}. Perfect traffic and weather conditions are assumed, and the traveling speed of the simulated vehicle is set at 30 km/h. As a demonstration of the proposed framework, a limited design space is explored, where the number of CPU cores (from 1 to 28, incremented by one), CPU frequency (1.0, 1.2, 1.5, 1.8, and 2.1 GHz), and LiDAR data publishing frequency (7 and 14 Hz) are varied. \revision{While real-world autonomous driving stacks expose many more hardware and software knobs (e.g., GPU accelerators, memory/QoS settings, multi-sensor suites, and alternative planning algorithms), our work focuses on the above three parameters as a proof-of-concept to isolate and study the behavior of the LLM-based DSE loop under a controlled and fully enumerable design space.} The confined search space allows us to observe how the DSE Agent explores the space and to quantitatively estimate the quality of the identified DSE solutions in the following subsections. 

\begin{figure}[h!]
    \centering
    \begin{subfigure}[b]{0.23\textwidth} 
        \centering
        \includegraphics[width=.9\textwidth]{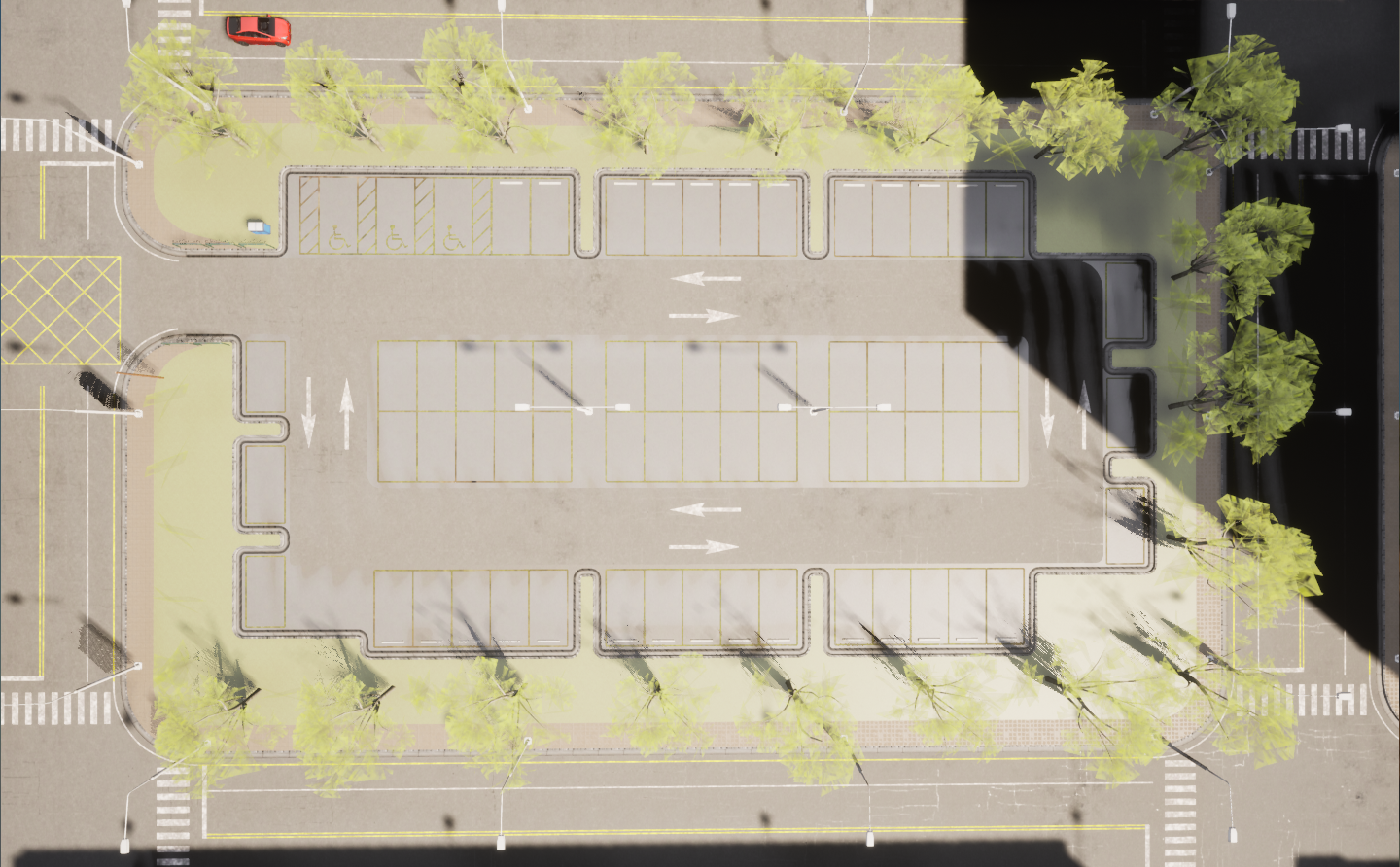} 
        \caption{Parking space map.}
        \label{fig:perf-data-map}
    \end{subfigure}
    \begin{subfigure}[b]{0.23\textwidth} 
        \centering
        \includegraphics[width=.9\textwidth]{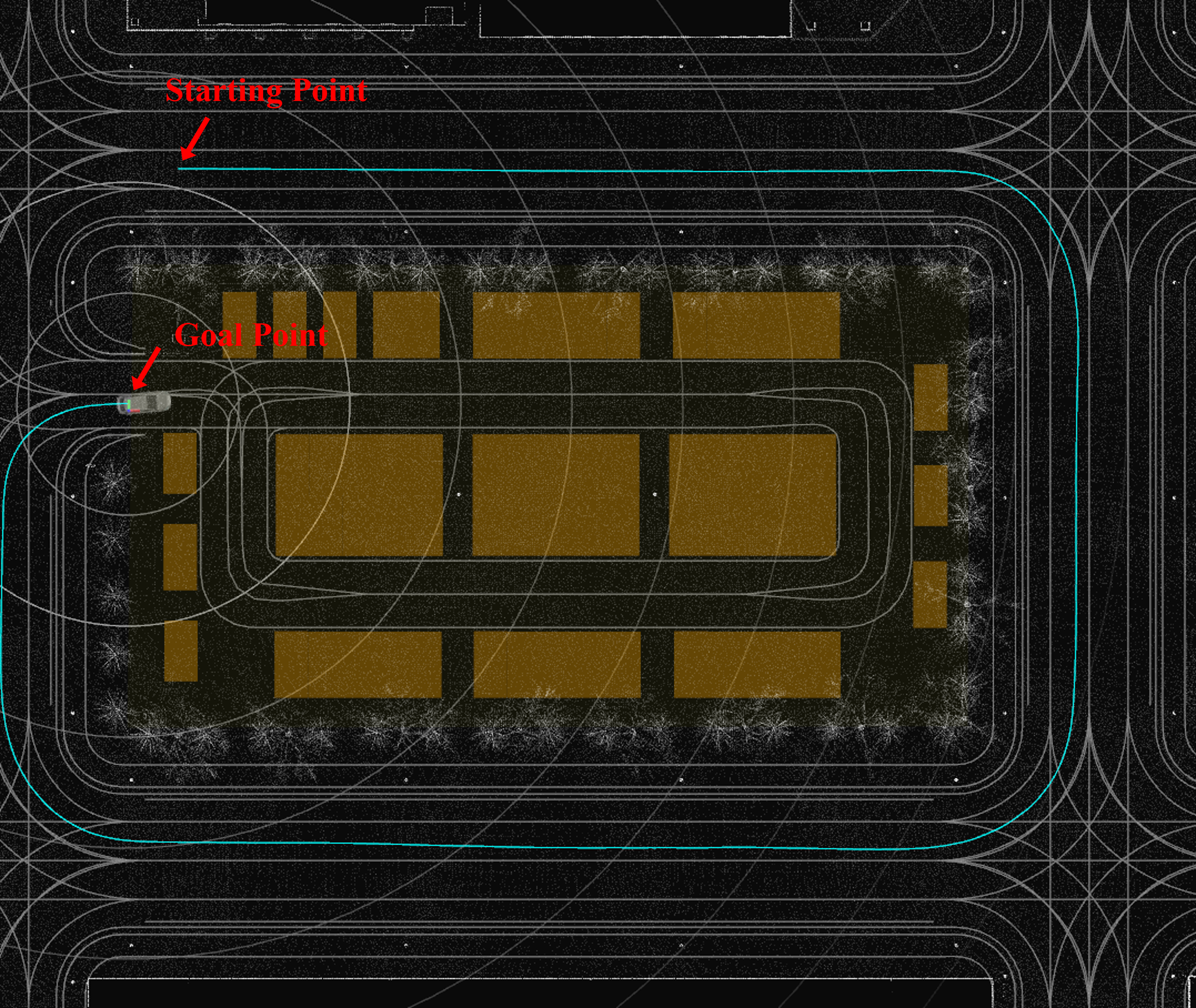} 
        \caption{Lane driving scenario.}
        \label{fig:perf-data-ld}
    \end{subfigure}
    \hfill 
    \begin{subfigure}[b]{0.23\textwidth}
        \centering
        \includegraphics[width=.9\textwidth]{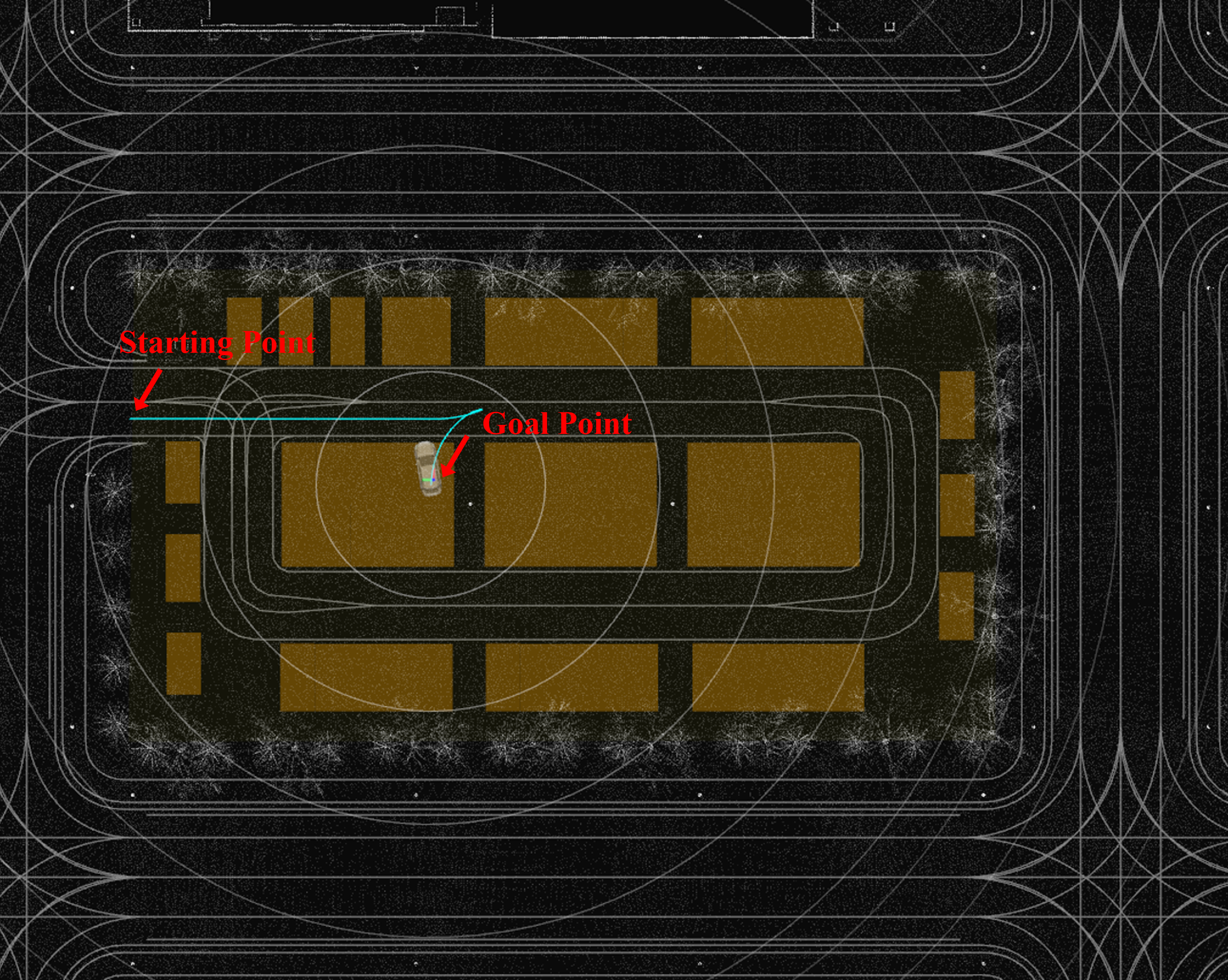}
        \caption{Automated valet parking scenario.}
        \label{fig:perf-data-avp}
    \end{subfigure}
    \caption{A parking space map used for the autonomous driving simulation (a), and the trajectories of the simulated vehicle results for lane driving scenario (b) and automated valet parking scenario (c).}
    \label{fig:perf-data-deciphering}
\end{figure}

As an application constraint, the time taken by the robotaxi to travel from the start position to the goal position is set as the low-bound of the navigation time, which is set to 400 seconds in this case study. Besides, the control command issue rate is also considered as another application constraint, which is defined as the frequency of the control commands issued by the robotaxi system to the simulated vehicle and should be larger than one Hz to ensure the safety of the self-driving vehicle~\cite{Becker2020}. 
The DSE task is performed on a parking space map, as shown in \figurename~\ref{fig:perf-data-map}, and the result of a successful simulation run (the vehicle trajectory) is illustrated in \figurename~\ref{fig:perf-data-ld}. The result of a successful automated valet parking simulation run is shown in \figurename~\ref{fig:perf-data-avp}, where the vehicle is controlled to park itself in a designated parking slot. 

\begin{figure}[bth!]
\centering
\includegraphics[width=0.7\columnwidth]{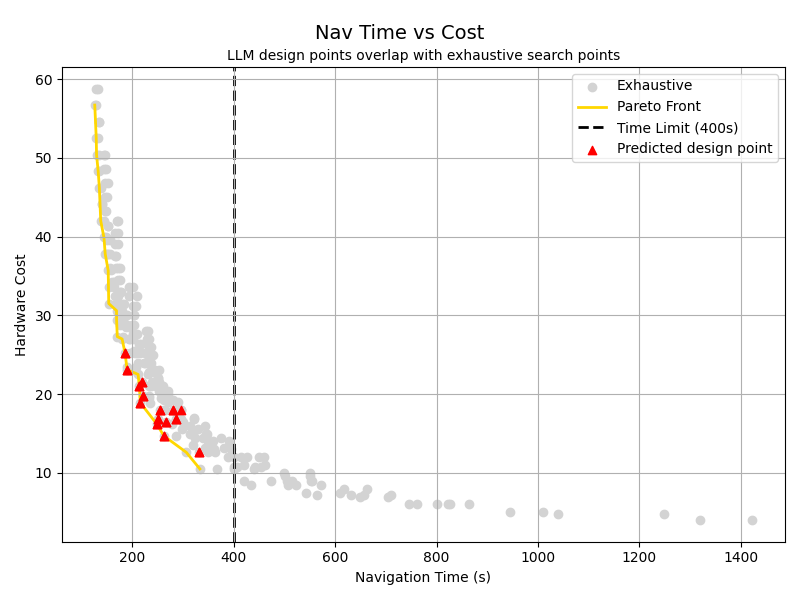}
  \caption{The search results identified by the exhaustive search method for the DSE task of the lane driving scenario.}
\label{fig:exhaustive_search}
\end{figure}

The results of the DSE task are plotted in \figurename~\ref{fig:exhaustive_search}, where the x-axis is the navigation time, and the y-axis is the hardware cost (the number of cores and the CPU frequency). A vertical dashed line at the navigation time of 400 seconds represents the time limit for the application; a design point with the time larger than 400 cannot be acceptable as a qualified solution. The gray points represent all of the possible designs explored by the exhaustive search method, and the yellow curve indicates the \emph{Pareto Front} solutions. 
The red triangle design points denote the candidate solutions discovered by our framework. There are some design points which overlap the design points on the pareto front found by the exhaustive search method, which indicates that our framework can effectively explore the design space and identify pareto front solutions.

\subsection{Search Efficiency Evaluation} \label{sec:sefficiency}
With the DSE results shown in \figurename~\ref{fig:exhaustive_search}, we further evaluate the search efficiency of our LLM-based DSE framework by comparing it with a traditional methodology, the genetic algorithm (GA)~\cite{GA}. GA is a widely used heuristic technique that emulates the process of natural selection, such as crossover and mutation, to find optimal solutions. Similarly, our LLM based framework adopts a sample-and-learn based exploration strategy. Such an iterative refinement scheme makes GA a suitable baseline for comparison and to observe the intelligence of the DSE Agent under the same exploration budget. In this experiment, the GA is integrated into the DSE Agent as a replacement for the LLM to predict good design points. The initial design points are selected randomly, and our framework begins with a hardware cost of 18 (12 cores, 1.5 GHz, and a 14 Hz LiDAR publishing frequency), while the GA starts with a hardware cost of 16.8 (14 cores, 1.2 GHz, and a 7 Hz LiDAR publishing frequency). 

\begin{figure}[hbt!]
\centering
\includegraphics[width=0.7\columnwidth]{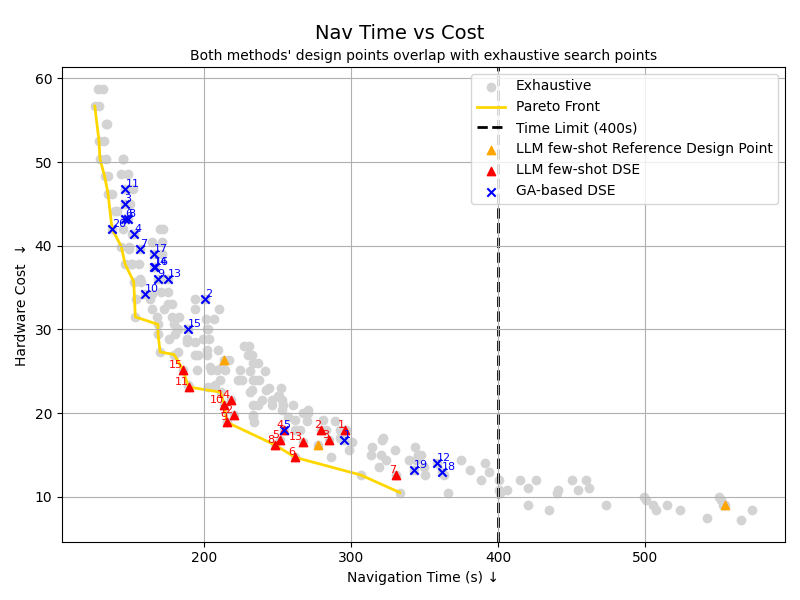}
\caption{Design points identified by our LLM- and GA-based approaches under the same exploration budget.}
\label{fig:searchquality}
\end{figure}

\figurename~\ref{fig:searchquality} shows the design points identified by the LLM- and GA-based approaches, where the x-axis is the navigation time (less than 400 seconds), and \revision{the y-axis is labeled hardware cost in these figures corresponds to $C_{\text{HW}} = N_{\text{cores}} \times f_{\text{core}}$ and should be interpreted as a relative CPU capacity-based cost proxy rather than an absolute monetary value.} Given an exploration budget of 15 iterations, the LLM-based DSE Agent identifies 6 design points on the Pareto front, whereas the GA-based approach identifies 0 Pareto front design point. Moreover, the LLM-based approach concentrates on design points with navigation time under 250 seconds (left side of the figure), while the GA-based approach explores the design space broadly across the valid navigation time range. As a few-shot prompting technique is adopted, the LLM-based approach is tasked with finding cost-efficient design points, which are expected to lie near the bottom-left corner of the figure.Moreover, as shown in \figurename~\ref{fig:searchquality}, we extended the experiment to determine when the GA-based DSE first reached a Pareto-front solution. The results show that the GA-based DSE achieved its first Pareto-front solution at iteration 20.

This shows that our LLM-based DSE framework can effectively explore the design space and identify high-quality solutions with fewer iterations compared to the traditional method. The impact of different prompting techniques on the search efficiency is further analyzed. However, due to the page length limitation, the related content is omitted. 

\section{Conclusion} \label{conclusion}

This paper presents an LLM-based DSE framework to address the challenges of software complexity and environmental variability when designing autonomous driving systems. By integrating multi-modal large language models, the framework’s Performance Deciphering Agent is able to interpret both visual and textual simulation results to understand the runtime behaviors, identify potential performance bottlenecks, and guide the DSE Agent toward cost-efficient design points. The use of the multi-modal LLMs is the key to the automation for the system design without human intervention for the analysis of the simulation results. As a case study, our prototype framework is applied on the robotaxi application, and our LLM-based approach outperforms a genetic algorithm baseline under the same exploration budget by discovering more Pareto front solutions. These solutions not only meet performance constraints but also offer cost-efficient configurations with shorter navigation times. The experimental results demonstrate that LLM intelligence can effectively correlate design parameters with design objectives, which leads to a more efficient exploration of the design space.

\printbibliography










\end{document}